\documentclass[conference]{IEEEtran}
\IEEEoverridecommandlockouts
\usepackage{cite}
\usepackage{amsmath,amssymb,amsfonts}
\usepackage{algorithmic}
\usepackage{graphicx}
\usepackage{textcomp}
\usepackage{xcolor}
\usepackage{tabularx}  
\usepackage{array}     
\usepackage{booktabs}  
\usepackage{caption}   
\usepackage{bm}       
\usepackage{float}
\usepackage{makecell}
\usepackage{multicol}
\usepackage{float}

\def\BibTeX{{\rm B\kern-.05em{\sc i\kern-.025em b}\kern-.08em
    T\kern-.1667em\lower.7ex\hbox{E}\kern-.125emX}}
\begin{document}

\title{A PID-Controlled Non-Negative Tensor Factorization Model for Analyzing Missing Data in NILM\\
}

\author{\IEEEauthorblockN{1\textsuperscript{st} Dengyu Shi}
\IEEEauthorblockA{\textit{School of Computer Science and Information)} \\
\textit{Southwest University}\\
Chongqing, China \\
tendyu@email.swu.edu.cn}
}

\maketitle

\begin{abstract}
With the growing demand for energy and increased environmental awareness, Non-Intrusive Load Monitoring (NILM) has become an essential tool in smart grid and energy management. By analyzing total power load data, NILM infers the energy usage of individual appliances without the need for separate sensors, enabling real-time monitoring from a few locations. This approach helps users understand consumption patterns, enhance energy efficiency, and detect anomalies for effective energy management. However, NILM datasets often suffer from issues such as sensor failures and data loss, compromising data integrity, thereby impacting subsequent analysis and applications. Traditional imputation methods, such as linear interpolation and matrix factorization, struggle with nonlinear relationships and are sensitive to sparse data, resulting in information loss. To address these challenges, this paper proposes a Proportional-Integral-Derivative (PID) Controlled Non-Negative Latent Factorization of Tensor (PNLF) model, which dynamically adjusts parameter gradients to improve convergence, stability, and accuracy. Experimental results show that the PNLF model significantly outperforms state-of-the-art tensor completion models in both accuracy and efficiency. By addressing data loss issues, this study enhances load disaggregation precision and optimizes energy management, providing reliable data support for smart grid applications and policy formulation.
\end{abstract}


\section{Introduction}

Non-Intrusive Load Monitoring (NILM) technology is widely used to estimate the energy consumption of individual appliances within buildings, providing detailed insights into electrical devices without the need for direct connections to each device. NILM technology plays a crucial role in optimizing energy utilization and achieving efficient energy use. However, potential issues such as sensor or smart meter faults, communication blockages, transmission delays, or network failures may lead to data loss, which impacts the accuracy of load monitoring and load disaggregation. Therefore, addressing data loss in NILM is essential to ensure the deployment of advanced applications like demand response \cite{paper41}, user behavior analysis \cite{paper42}, and energy consumption analysis \cite{paper43} on a high-quality data foundation.

Currently, there is no universal strategy for addressing data loss in NILM \cite{paper44}. In cases of low data loss and stable variations, common interpolation methods \cite{paper45} and K-nearest neighbors \cite{paper46} are typically employed. However, in scenarios with higher data loss, these methods may not be suitable for data imputation. Traditional data imputation methods mainly include regression analysis and matrix decomposition. Regression methods \cite{paper47} often assume linear relationships among variables, limiting their ability to capture nonlinear relationships. Principal Component Analysis (PCA) \cite{paper48}, while widely used, may perform poorly when the data exhibits nonlinear relationships. Factor decomposition \cite{paper49,paper50} is advantageous for understanding data structures and relationships, but the selection of the number of factors may lead to information loss or overfitting.

As data evolves, it can exhibit diverse patterns, and matrix decomposition may encounter difficulties in capturing the complex characteristics of the data. Tensor completion methods have been extensively researched. These methods integrate multiple sources of information into tensor structures, effectively addressing irregular missing patterns and achieving high-precision recovery of missing data. Initially, Liu et al. \cite{paper51} defined tensor nuclear norm as a combination of matrix nuclear norms obtained by unfolding the tensor along its modes. Kilmer et al. \cite{paper52} proposed the tensor singular value decomposition (t-SVD) method and subsequently defined a new tensor nuclear norm (TNN) in their later research \cite{paper53}, representing it as the sum of absolute values of all elements in the core tensor. t-SVD has since been widely applied in tasks such as video and image processing \cite{paper54, paper55}. Despite their effectiveness, these methods face challenges in terms of computational efficiency and memory requirements due to the necessity of singular value decomposition \cite{paper56}.

In recent years, Latent Factorization of tensor (LFT) models have gained widespread attention for their exceptional performance. For example, Luo et al. enhanced the robustness of the non-negative LFT model by introducing bias terms \cite{paper10}. Acar et al.'s CP-WOPT algorithm \cite{paper57} effectively transformed the CANDECOMP/PARAFAC (CP) decomposition problem into a weighted least squares approach, using first-order optimization techniques to recover the underlying structure from known data while ignoring missing values. Zhang et al.'s Non-negative Tensor Factorization (NTF) model \cite{paper58} employed a multiplicative update rule to iteratively ensure non-negativity in the factor matrices. Additionally, Wu et al. proposed the Fused CP (FCP) decomposition model \cite{paper59}, which integrates various priors, such as low rank, sparsity, manifold information, and smoothness, to enhance tensor completion performance.

Despite the advancements made by existing LFT models in addressing data loss issues, limitations persist, including slow convergence speeds and insufficient handling of nonlinear features. To overcome these challenges, this paper proposes a Proportional-Integral-Derivative (PID) Controlled Non-Negative Latent Factorization of Tensor (PNLF) model, which offers the following key contributions:

1) The proposed model integrates a PID controller, which differs from the approach in \cite{paper12} where PID is used to control instance errors. In this work, the PID dynamically adjusts the gradients to accelerate the training process. Additionally, a Sigmoid function is employed to ensure data non-negativity, while nonlinear components are used to effectively capture complex data features;

2) The paper provides detailed algorithm design and model analysis, offering specific guidance for researchers applying the PNLF model in data imputation tasks.

Experimental results demonstrate that the proposed PNLF model significantly outperforms existing state-of-the-art models in terms of efficiency and accuracy for recovering missing data in NILM. The remainder of this paper is organized as follows: Section \ref{sec2} introduces relevant background knowledge, Section \ref{sec3} presents the proposed method, Section \ref{sec4} discusses the experimental results, and Section \ref{sec5} discusses the findings of the paper.

\section{Preliminaries}
\label{sec2}
\subsection{Symbols Appointment}

Table ~\ref{tab1} provides a detailed description of the symbols used in this paper.

\begin{table}[!htbp]
\caption{Adopted Symbols and Their Description. }
\centering
\begin{tabular}{cc}
\toprule[2pt] 
Symbol & Description \\
\midrule 
$I,J,K$ & Three entity sets  \\
$ \bm{\mathit{Y}}$ & Three-order target tensor \\
$ \bm{\mathit{X}}$ & Rank-one tensor \\
$ \bm{\mathit{\hat{Y}}}$ & approximation to $\bm{\mathit{Y}}$\\
$ \bm{\mathit{X}}_{r}$ & The $R$-th rank-one tensor summed to form tensor $ \bm{\mathit{\hat{Y}}}$
\\
$ y_{ijk},x_{ijk},\hat{y}_{ijk}$ & A single element in$ \bm{\mathit{Y}},\bm{\mathit{X}}$ and $ \bm{\mathit{\hat{Y}}}$ \\
$ R $ &\makecell[c]{ Rank of $\bm{\mathit{\hat{Y}}},$ dimension of the latent feature space }\\
$ U,O,M$ & latent feature matrices (LFs)\\
$ U_{,r},O_{,r},M_{,r} $ & The $r$-th latent feature vectors in $U, O$ and $M$\\
$ u_{ir}, o_{jr}, m_{kr} $ & Single element in $U, O$ and $M$\\

$ \lambda $ & Regularization coefficient\\

$ \eta $ & Learning rate\\

$ \circ  $ & Outer product of two vectors\\
$ \left | \cdot  \right |  $ & Cardinality of a set\\
$ \left \| \cdot  \right \| _F $ & Frobenius norm of an enclosed tensort\\
$ \Lambda  $  & Known sets of tensor $ \bm{\mathit{Y}}$\\
$ \Phi,\Psi,\Omega $ & Training, validation and testing sets from  $ \Lambda  $\\

$ C_{P},C_{I},C_{D} $ &\makecell[c]{ Controlling coefficients of 
proportional, integral \\ and
derivative terms}\\
$I_{(U)}, I_{(O)} , I_{(M)}$ &\makecell[c]{ The auxiliary matrices store  the cumulative \\ updates  of  $U$, $O$ and $M$ }\\

$D_{(U)}, D_{(O)} , D_{(M)}$ &\makecell[c]{ The auxiliary matrices store the previous \\ updates of $U$, $O$ and $M$}\\

\bottomrule[2pt] 
\end{tabular}
\label{table1}
\end{table}

\subsection{Tensorization of NILM Data}

During the modeling process, the different dimensions of NILM data (dates, time steps, and devices) may have inherent correlations. To better capture these relationships, we assume that the data can be represented as a low-rank tensor. This means that although the data appear high-dimensional on the surface, they can be approximated by a few underlying patterns.

\begin{figure}[htbp]
\centering
\includegraphics[width=0.5 \textwidth]{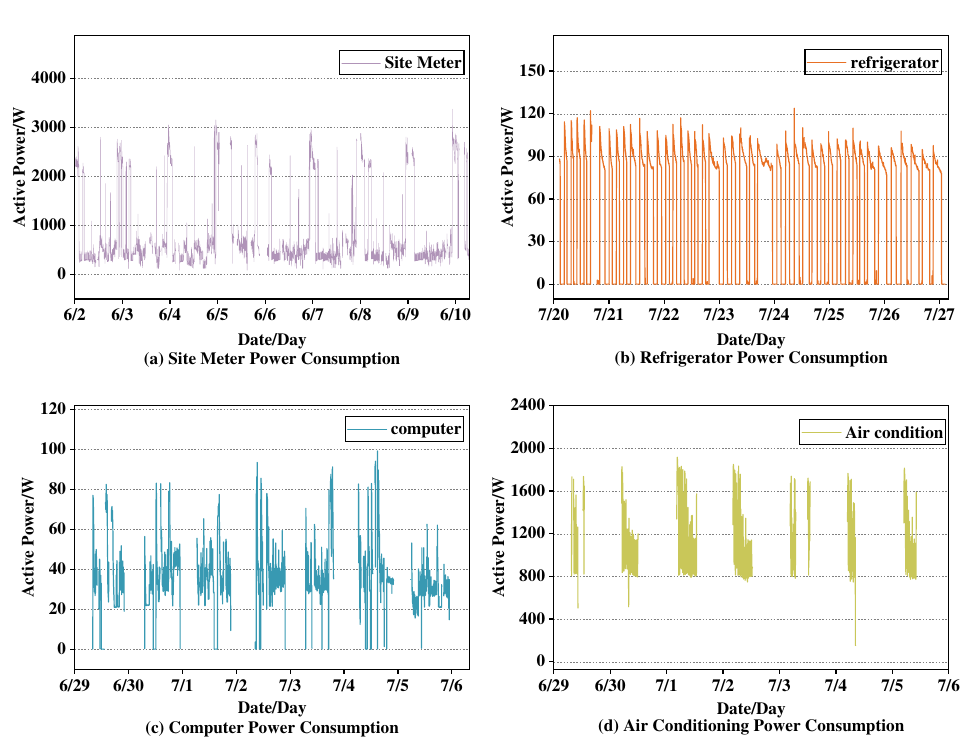}
\caption{Power variations of three appliances over a period of dates. The data is sourced from the IAWE dataset \cite{paper30}.}
\label{fig1}
\end{figure}

\begin{itemize}

\item	\textbf{Date Dimension:} This dimension distinguishes between different dates, allowing us to identify consumption differences between weekdays and weekends, which is crucial for understanding how energy usage varies by day of the week;
\item	\textbf{Time Steps Dimension:} This dimension captures the details of energy consumption at different times of the day, revealing differences between morning and evening usage as well as periodic trends across days. For example, some devices may exhibit cyclical energy consumption patterns, as shown in Figure~\ref{fig1}(a) and Figure~\ref{fig1}(b); 
\item	\textbf{Meters Dimension}: This dimension represents the various devices in the monitored environment, allowing us to analyze relationships between them. For instance, the usage of some devices (such as computers) at specific times may be correlated with the usage of others (such as air conditioners), as shown in Figure~\ref{fig1}(c) and Figure~\ref{fig1}(d).

\end{itemize}

\begin{figure*}[htbp]
\centering
\includegraphics[width=1 \textwidth]{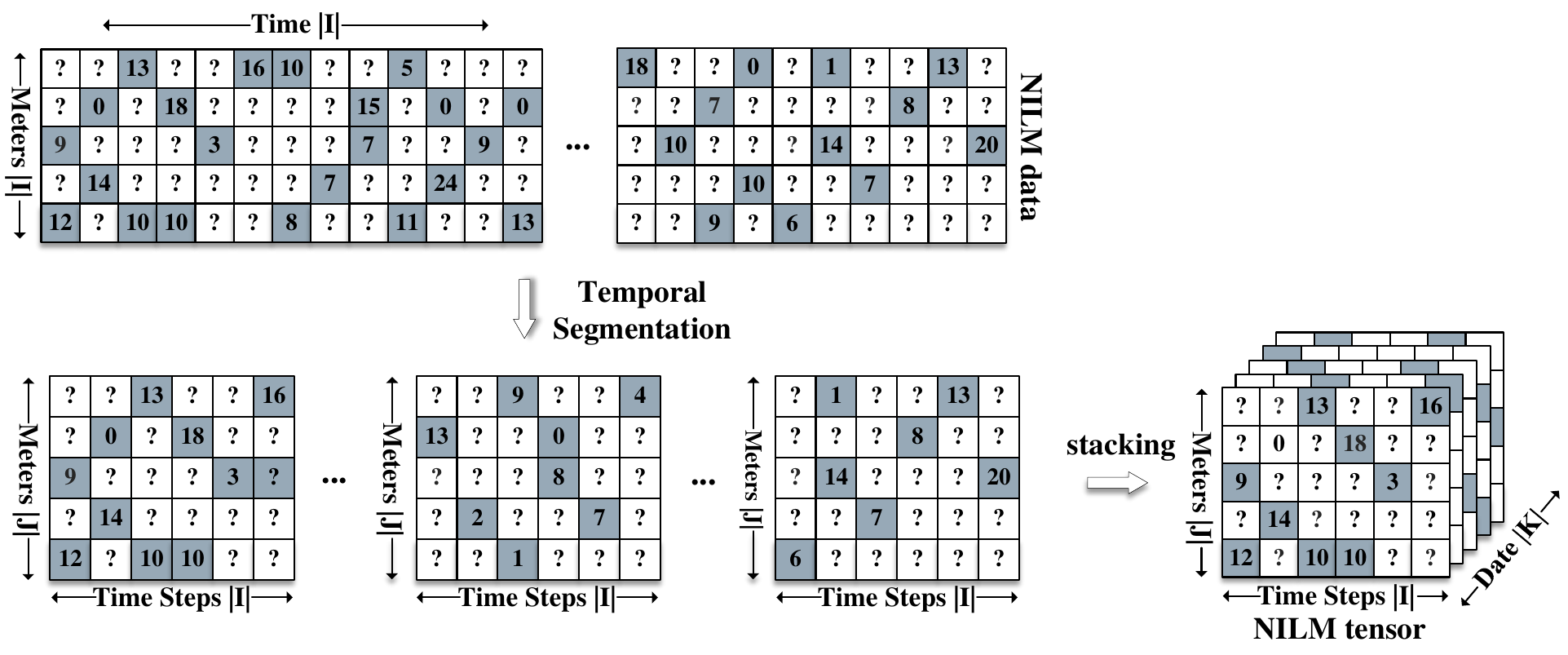}
\caption{Tensorization Process: Transformation of NILM Data.\label{fig2}}
\end{figure*}

\textbf{\textit{Definition 1}}\textit{(NILM Tensor)}: As shown in Figure~\ref{fig2}, a three-dimensional tensor $\bm{\mathit{Y}}^{|I|\times|J|\times|K|}$ is utilized in the process of converting NILM data into a NILM tensor. Here, dimensions $I$, $J$, and $K$ correspond to specific time steps, meters, and date, respectively. Each element $y_{ijk}$ in the tensor $\bm{\mathit{Y}}$ represents the power consumption of a specific meter $j$ at a particular time $i$ on a specific date $k$.

\subsection{LFT}

\textbf{\textit{Definition 2}}\textit{(rank-one tensor)}. For a three-dimensional tensor $\bm{\mathit{X}}^{|I|\times|J|\times|K|}$, its elements $x_{ijk}$  can be represented as the product of three scalars, i.e., $x_{ijk}=a_{i}b_{j}c_{k}$ , where $a$,  $b$, and $c$  are vectors of length $I$,  $J$, and  $K$, respectively, and  $i$, $j$, $k$ are the corresponding indices. When $\bm{\mathit{X}}$ can be expressed as the outer product of these three vectors, it is referred to as a rank-one tensor, as follows:

\begin{equation}
\bm{\mathit{X}} = a \circ b \circ c,
\label{eq1}
\end{equation}

\textbf{\textit{Definition 3}} \textit{(CP decomposition)}: CP decomposition transforms high-dimensional tensors into low-dimensional factor matrices, effectively reducing the dimensionality of the data \cite{paper23}. CP decomposition can be represented as follows: Given a tensor, it can be decomposed into the sum of multiple rank-one tensors. Specifically, for a three-dimensional tensor $\bm{\mathit{\hat{Y}}}$, it can be expressed as the sum of $R$ rank-one tensors:

\begin{equation}
    \bm{\mathit{\hat{Y}}}= \sum\limits_{r = 1}^R {{\bm{\mathit{X}}_{r}}}  = \sum\limits_{r = 1}^R {{U_{,r}} \circ {O_{,r}} \circ {M_{,r}}},
    \label{eq2}
\end{equation}
where each element $\hat y_{ijk}$ in $ \bm{\mathit{\hat{Y}}}$ can be represented as:

\begin{equation}
    {\hat y_{ijk}} = \sum\limits_{r = 1}^R {{u_{ir}}{o_{jr}}{m_{kr}}}.
    \label{eq3}
\end{equation}

\begin{figure*}[htbp]
\centering
\includegraphics[width=1 \textwidth]{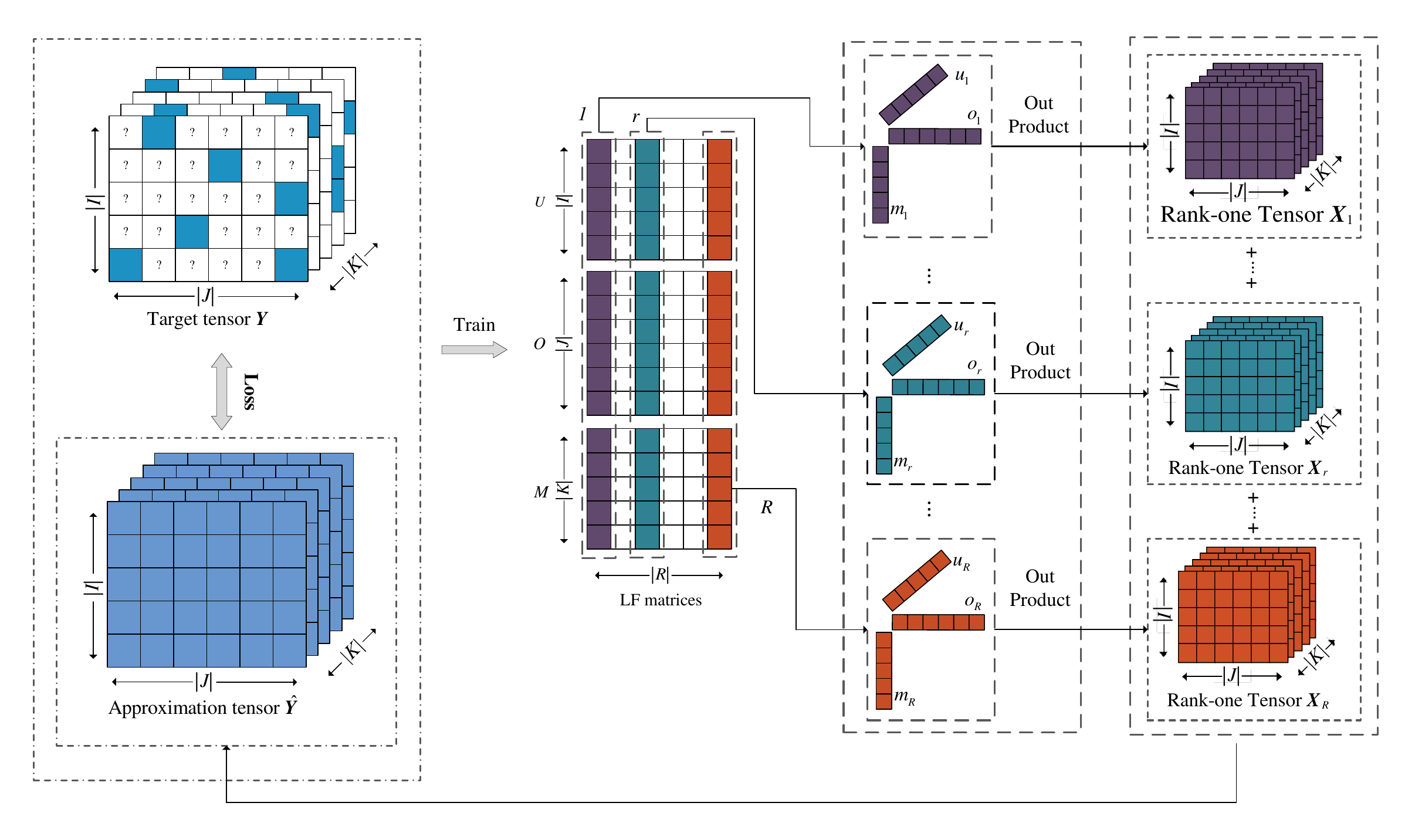}
\caption{Latent factorization of a target tensor \bm{\mathit{Y}}. \label{fig3}}
\end{figure*}

As shown in Figure~\ref{fig3}, the CPD-based LFT model requires obtaining U, O, and M to construct the approximation $\bm{\mathit{\hat{Y}}}$ of $\bm{\mathit{Y}}$. Figure~\ref{fig3} illustrates the process of decomposing a three-dimensional tensor into three latent feature (LF) matrices. To obtain the desired LFs, the Euclidean distance is utilized to quantify the difference between $\bm{\mathit{\hat{Y}}}$ and $\bm{\mathit{Y}}$. The objective function $f$ is given as follows:
\begin{equation}
    f = \frac{1}{2}\left\| {{\bm{\mathit{Y}}} - {\bm{\mathit{\hat{Y}}}}} \right\|_F^2.
    \label{eq4}
\end{equation}

Due to the limited number of known entries in $\bm{\mathit{Y}}$, $f$ is solely defined on $\Lambda$ and further represented as:

\begin{align}
    f &= \frac{1}{2}{\sum\limits_{{y_{ijk}} \in \Lambda } {\left( {{y_{ijk}} - {{\hat y}_{ijk}}} \right)} ^2} \nonumber\\
    &= \frac{1}{2}{\sum\limits_{{y_{ijk}} \in \Lambda } {\left( {{y_{ijk}} - \sum\limits_{r = 1}^R {{u_{ir}}{o_{jr}}{m_{kr}}} } \right)} ^2}.
    \label{eq5}
\end{align}

Given the ill-posed nature of the aforementioned equation, Tikhonov regularization is introduced, as suggested by \cite{paper10,paper11,paper12,paper24}, to mitigate overfitting and enhance model stability. $f$ is represented as:

\begin{equation}
    f = \frac{1}{2}\sum\limits_{{y_{ijk}} \in \Lambda } {\left( \begin{array}{l}
{\left( {{y_{ijk}} - \sum\limits_{r = 1}^R {{u_{ir}}{o_{jr}}{m_{kr}}} } \right)^2} \\
  + \lambda \sum\limits_{r = 1}^R {\left( {u_{ir}^2 + o_{jr}^2 + m_{kr}^2} \right)} 
\end{array} \right)} .
\label{eq6}
\end{equation}

where $\hat{y}_{ijk}$ representing the approximation term corresponding to each instance $y_{ijk}$.

\subsection{PID Controller}

The PID controller utilizes current, past, and future instance error information to regulate feedback systems \cite{paper9,paper12,paper25,paper26}. It adjusts instantaneous error based on proportional, integral, and derivative terms. The proportional term scales with the current error, the integral term accounts for the cumulative past error, and the derivative term considers the rate of change of the error. At time point $t$, the discrete PID controller constructs the adjusted instantaneous error as:

\begin{equation}
    {\tilde e^t} = {C_P}{e^t} + {C_I}\sum\limits_{i = 1}^t {{e^i}}  + {C_D}({e^t} - {e^{t - 1}}),
    \label{eq7}
\end{equation}
where ${e^t}$ represent the instantaneous errors at time point $t$, while  ${C_P}$, ${C_I}$ and ${C_D}$ are the gain coefficients for the proportional, integral and derivative terms. Note that the error ${\tilde e^t}$, defined as the difference between the desired and actual output, shares the same essence as the gradient used in machine learning optimization. In this paper, PID theory is explored as a novel optimizer, inheriting the excellent characteristics of PID controllers while retaining simplicity and efficiency.

\section{Methods}
\label{sec3}

\subsection{Standard SGD-Based Non-Negative LFT Model}

Although CP decomposition typically assumes that factor matrices are linear, complex nonlinear relationships are often observed in real-world data. By introducing the sigmoid function, nonlinear transformations can be incorporated, enhancing the model's expressive power and better capturing the intricate relationships among the data \cite{paper27,paper28}. Additionally, considering the non-negative nature of NILM data, the sigmoid function confines its output to the range (0, 1), ensuring that the elements of the factor matrices remain non-negative. This ensures better adaptation to the characteristics of real-world data, thereby improving the accuracy and generalization capability of the model. The sigmoid function is defined as follows:
\begin{equation}
    \varphi (x) = \frac{1}{{1 + {e^{-x}}}}.
    \label{eq8}
\end{equation}

Combining  (\ref{eq6})  and (\ref{eq8}), the objective optimization function is given as:

\begin{equation}
    \varepsilon=\frac{1}{2}\sum\limits_{{y_{ijk}} \in \Lambda } {\left( \begin{array}{l}
    {\left( {{y_{ijk}} - \sum\limits_{r = 1}^R {\varphi \left( {{u_{ir}}} \right)\varphi \left( {{o_{jr}}} \right)\left( {{m_{kr}}} \right)} } \right)^2} + \\
    \lambda \sum\limits_{r = 1}^R {\left( {\varphi {{\left( {{u_{ir}}} \right)}^2} + \varphi {{\left( {{o_{jr}}} \right)}^2} + \varphi {{\left( {{m_{kr}}} \right)}^2}} \right)} 
    \end{array} \right)} .
\end{equation}

Based on prior research  \cite{paper7,paper8,paper9,paper10,paper11,paper12,paper29}, in LF analysis, the Stochastic Gradient Descent (SGD) algorithm demon-strates high computational efficiency and ease of implementation. Utilizing SGD, the update rules for LFs are given as:

\begin{equation}
\begin{aligned}
    &\text{arg} \min_{S,D,T} \varepsilon \overset{SGD}{\Longrightarrow} \forall i \in I, j \in J, k \in K, r \in \{1, \ldots, R\} \\
    &\begin{cases}
        u_{ir}^{t} \leftarrow u_{ir}^{t-1} - \eta \frac{\partial f_{ijk}^{t-1}}{\partial u_{ir}^{t-1}} \\[0.8ex]
        o_{jr}^{t} \leftarrow o_{jr}^{t-1} - \eta \frac{\partial f_{ijk}^{t-1}}{\partial o_{jr}^{t-1}} \\[0.8ex]
        m_{kr}^{t} \leftarrow m_{kr}^{t-1} - \eta \frac{\partial f_{ijk}^{t-1}}{\partial m_{kr}^{t-1}}
    \end{cases}
    .
\end{aligned}
\label{eq7}
\end{equation}

The detailed calculations for the gradients are given as:

\begin{equation}
\left\{
\begin{array}{l}
\frac{{\partial f_{ijk}}}{{\partial u_{ir}}} = \lambda \varphi \left( {{u_{ir}}} \right) \left( 1 - \varphi \left( {{u_{ir}}} \right) \right) {u_{ir}} - \\[0.2ex] 
\varepsilon \varphi \left( {{u_{ir}}} \right) \left( 1 - \varphi \left( {{u_{ir}}} \right) \right) \varphi \left( {{o_{jr}}} \right) \varphi \left( {{m_{kr}}} \right) \\[2ex] 
\frac{{\partial f_{ijk}}}{{\partial o_{jr}}} = \lambda \varphi \left( {{o_{jr}}} \right) \left( 1 - \varphi \left( {{o_{jr}}} \right) \right) {o_{jr}} - \\[0.2ex] 
\varepsilon \varphi \left( {{o_{jr}}} \right) \left( 1 - \varphi \left( {{o_{jr}}} \right) \right) \varphi \left( {{u_{ir}}} \right) \varphi \left( {{m_{kr}}} \right) \\[2ex] 
\frac{{\partial f_{ijk}}}{{\partial m_{kr}}} = \lambda \varphi \left( {{m_{kr}}} \right) \left( 1 - \varphi \left( {{m_{kr}}} \right) \right) {m_{kr}} - \\[0.2ex] 
\varepsilon \varphi \left( {{m_{kr}}} \right) \left( 1 - \varphi \left( {{m_{kr}}} \right) \right) \varphi \left( {{u_{ir}}} \right) \varphi \left( {{o_{jr}}} \right)
\end{array},
\right.
\end{equation}
where 
$\varepsilon  = {y_{ijk}} - {\textstyle\sum_{r = 1}^{R}} {\varphi ({u_{ir}})\varphi ({o_{jr}})\varphi ({m_{kr}})}  = {y_{ijk}} - {\hat y_{ijk}}$ represents the instance loss on each individual training instance.

\subsection{PID Control in Non-Negative LFT Model}
\subsubsection{SGD as a P Controller}
The parameter update rule of SGD from time $t$ to  $t+1$ is given by:
\begin{equation}
{\theta _{t + 1}} = {\theta _t} - \eta {{\partial {L_t}} \mathord{\left/
 {\vphantom {{\partial {L_t}} {\partial {\theta _t}}}} \right.
 \kern-\nulldelimiterspace} {\partial {\theta _t}}},
\end{equation}
where  $\eta $ controlling the step size of each parameter update, ${\theta _t}$  and  ${L_t}$ represent the parameter value and objective function value at the $t$-th iteration, respectively. Viewing the gradient   
${{\partial {L_t}} \mathord{\left/
 {\vphantom {{\partial {L_t}} {\partial {\theta _t}}}} \right.
 \kern-\nulldelimiterspace} {\partial {\theta _t}}}$ as the error ${e^t}$   and comparing (11) to the PID controller in (7),  $\eta$ functions similarly to the proportional gain  
 $C_{P}$ , adjusting the update based on the current gradient. Thus, SGD can be regarded as a P controller with  ${C_P} = \eta $ .

 \subsubsection{SGD-Momentum as a PI Controller}
SGD-Momentum accelerates convergence by accumulating historical gradient information. The parameter update rule is as follows:
\begin{equation}
    \left\{ \begin{array}{l}
    {V_{t + 1}} = \alpha {V_t} - \eta \frac{{\partial {L_t}}}{{\partial {\theta _t}}}{\rm{ }}\\
    {\theta _{t + 1}} = {\theta _t} + {V_{t + 1}}
    \end{array} \right.,
    \label{eq12}
\end{equation}
 where ${V_{t + 1}}$  represents the accumulation of past gradients. With some mathematical transformations \cite{paper26},  (\ref{eq12}) can be rewritten as:

 \begin{equation}
     {\theta _{t + 1}} = {\theta _t} - \eta {{{\partial {L_t}} \mathord{\left/
 {\vphantom {{\partial {L_t}} {\partial \theta }}} \right.
 \kern-\nulldelimiterspace} {\partial \theta }}_t} - \eta \sum\limits_{i = 0}^t {\left( {{\alpha ^{t - i}}{{{{\partial {L_i}} \mathord{\left/
 {\vphantom {{\partial {L_i}} {\partial \theta }}} \right.
 \kern-\nulldelimiterspace} {\partial \theta }}}_i}} \right)} .
 \end{equation}
 It can be seen that the parameter update depends not only on the current gradient $\left({{{\partial {L_t}} \mathord{\left/
 {\vphantom {{\partial {L_t}} {\partial \theta }}} \right.
 \kern-\nulldelimiterspace} {\partial \theta }}_t} \right)$ but also on the accumulated sum of past gradients $\eta \sum\nolimits_{i = 0}^t {\left( {{\alpha ^{t - i}}{{{{\partial {L_i}} \mathord{\left/
 {\vphantom {{\partial {L_i}} {\partial \theta }}} \right.
 \kern-\nulldelimiterspace} {\partial \theta }}}_i}} \right)} $ . Unlike the integral term in a PI controller, it includes a decay factor, $\alpha$ , which helps reduce the influence of distant past gradients on the current update. Overall, SGD-Momentum can be considered a type of PI controller.

 \subsubsection{Overshoot Problem in SGD-Momentum}
 Overshoot occurs primarily due to the accumulation of historical gradients in momentum methods. During optimization, momentum methods accumulate past gradient information to accelerate convergence. However, when the model needs to change the optimization direction, the accumulated historical gradients can cause a lag in updating the model weights, leading to excessive updates that overshoot the target, resulting in overshoot phenomena  \cite{paper26,paper39}.

 The PID optimizer addresses overshoot by incorporating a derivative term. The PID optimizer is given as:
 \begin{equation}
     PID = PI + {C_D}\left( {{{\partial {L_n}} \mathord{\left/
 {\vphantom {{\partial {L_n}} {\partial {\theta _n}}}} \right.
 \kern-\nulldelimiterspace} {\partial {\theta _n}}} - {{\partial {L_{n - 1}}} \mathord{\left/
 {\vphantom {{\partial {L_{n - 1}}} {\partial {\theta _{n - 1}}}}} \right.
 \kern-\nulldelimiterspace} {\partial {\theta _{n - 1}}}}} \right),
 \end{equation}
 where $n$ denotes the current iteration number. The PID optimizer can detect rapid gradient changes by incorporating the term $\left({{{\partial {L_n}} \mathord{\left/
 {\vphantom {{\partial {L_n}} {\partial {\theta _n}}}} \right.
 \kern-\nulldelimiterspace} {\partial {\theta _n}}} - {{\partial {L_{n - 1}}} \mathord{\left/
 {\vphantom {{\partial {L_{n - 1}}} {\partial {\theta _{n - 1}}}}} \right.
 \kern-\nulldelimiterspace} {\partial {\theta _{n - 1}}}}} \right)$. When it detects that the gradient direction may need to reverse, the derivative term reduces the influence of historical gradients, preventing excessive updates and thereby mitigating overshoot.

 \subsubsection{PID-based Parameter Update}
 In this study, the PID controller is used to adjust parameter updates in the SGD algorithm for the LFT model, accelerating the update process. The PID-based SGD update method is as follows:

 \begin{equation}
     \left\{ \begin{array}{l}
{h_t} = \left( {1 - \alpha } \right){h_{t - 1}} + \alpha {{\partial {L_t}} \mathord{\left/
 {\vphantom {{\partial {L_t}} {\partial {\theta _t}}}} \right.
 \kern-\nulldelimiterspace} {\partial {\theta _t}}}\\
{\theta _{t + 1}} = {\theta _t} - \left( 
{{{\eta \partial {L_t}} \mathord{\left/
 {\vphantom {{\eta \partial {L_t}} {\partial {\theta _t}}}} \right.
 \kern-\nulldelimiterspace} {\partial {\theta _t}}} + {C_I}{h_t} + {C_D}
 
 \left( 
 {{{\partial {L_t}} \mathord{\left/
 {\vphantom {{\partial {L_t}} {\partial {\theta _t}}}} \right.
 \kern-\nulldelimiterspace} {\partial {\theta _t}}} - {{\partial {L_{t - 1}}} \mathord{\left/
 {\vphantom {{\partial {L_{t - 1}}} {\partial {\theta _{t - 1}}}}} \right.
 \kern-\nulldelimiterspace} {\partial {\theta _{t - 1}}}}} 
 
  \right)} \right)
\end{array} \right..
 \end{equation}

 By introducing a PID (Proportional-Integral-Derivative) optimizer to accelerate the model optimization pro-cess, the approach operates as follows:
 \begin{itemize}
\item	Proportional Term (P): Updates are based solely on the current gradient, similar to traditional SGD;
\item	Integral Term (I): Accumulates past gradient information to correct long-term errors in the model;
\item	Derivative Term (D): Uses the rate of change of the gradient (i.e., the gradient's derivative) to predict future gradient changes and make preemptive adjustments.
\end{itemize}
After combining equations (10) and (17), the updating rule for LF is given as follows:

\begin{equation}
    \left\{ \begin{array}{l}
I_{{u_{ir}}}^0 = 0,\\
{\rm{  }}u_{ir}^{t + 1} = u_{ir}^t - \left( {\eta \frac{{\partial {f^t}}}{{\partial u_{ir}^t}} + {C_I}I_{{u_{ir}}}^{t - 1} + {C_D}\left( {\frac{{\partial {f^t}}}{{\partial u_{ir}^t}} - \frac{{\partial {f^{t - 1}}}}{{\partial u_{ir}^{t - 1}}}} \right)} \right),
\\
{\rm{ }}I_{{u_{ir}}}^t = \left( {1 - \alpha } \right)I_{{u_{ir}}}^{t - 1} + \alpha \frac{{\partial {f^t}}}{{\partial u_{ir}^t}}.\\

\end{array} \right.
\end{equation}

\begin{equation}
    \left\{ \begin{array}{l}
I_{{o_{jr}}}^0 = 0,\\

{\rm{  }}o_{jr}^{t + 1} = o_{jr}^t - \left( {\eta \frac{{\partial {f^t}}}{{\partial o_{jr}^t}} + {C_I}I_{{o_{jr}}}^{t - 1} + {C_D}\left( {\frac{{\partial {f^t}}}{{\partial o_{jr}^t}} - \frac{{\partial {f^{t - 1}}}}{{\partial o_{jr}^{t - 1}}}} \right)} \right){\rm{, }}
\\
I_{{o_{jr}}}^t = \left( {1 - \alpha } \right)I_{{o_{jr}}}^{t - 1} + \alpha \frac{{\partial {f^t}}}{{\partial o_{jr}^t}}.
\end{array} \right.
\end{equation}

\begin{equation}
    \left\{ \begin{array}{l}

I_{{m_{kr}}}^0 = 0,
\\
{\rm{ }}m_{kr}^{t + 1} = m_{kr}^t - \left( {\eta \frac{{\partial {f^t}}}{{\partial m_{kr}^t}} + {C_I}I_{{m_{kr}}}^{t - 1} + {C_D}\left( {\frac{{\partial {f^t}}}{{\partial m_{kr}^t}} - \frac{{\partial {f^{t - 1}}}}{{\partial m_{kr}^{t - 1}}}} \right)} \right),
\\
{\rm{ }}I_{{m_{kr}}}^t = \left( {1 - \alpha } \right)I_{{m_{kr}}}^{t - 1} + \alpha \frac{{\partial {f^t}}}{{\partial m_{kr}^t}}.
\end{array} \right.
\end{equation}

Note that during the first round of parameter updates, both the integral and differential terms have values of 0. the auxiliary matrices ${I_{(U)}}$ , ${I_{(O)}}$, and ${I_{(M)}}$ are used to store the integral information of the LF matrices $U$, $O$ and $M$, while  ${D_{(U)}}$, ${D_{(O)}}$, and   ${D_{(M)}}$ store the previous updates. In the experiments, the decay factor  
$\alpha $ was set to 0.2. The PNLF model is now complete.

\begin{table*}[h]
\centering
\newcolumntype{C}{>{\raggedright\arraybackslash}X}

\newcolumntype{Y}{>{\centering\arraybackslash}p{6cm}} 
\begin{tabularx}{\textwidth}{CY}
\toprule[2pt] 
\textbf{Algorithm \  PNLF}  &    \\
\midrule 
\textbf{Input:} $\Lambda, R, I, J, K, \lambda, \eta, {C_I},{C_D}$ &   \\
\textbf{output:} $\varphi (U),\varphi (O),\varphi (M)$ &   \\
\midrule
\textbf{Operation} & \textbf{Cost}  \\
\midrule
\ \ 1:\textbf{Initialize} ${U^{|I| \times R}},{O^{|J| \times R}},{M^{|K| \times R}}$ with random numbers  in  the  range -3 to -2  & $\Theta \left( {\left( {\left| I \right| + \left| J \right| + \left| K \right|} \right) \times R} \right)$  \\
\ \ 2:\textbf{Initialize} $I^{|I|\times R}_{U}, I^{|J|\times R}_{O}, I^{|K|\times R}_{M} = 0$   & $\Theta \left( {\left( {\left| I \right| + \left| J \right| + \left| K \right|} \right) \times R} \right)$  \\
\ \ 3:\textbf{Initialize} $D^{|I|\times R}_{U}, D^{|J|\times R}_{O}, D^{|K|\times R}_{M} = 0$   & $\Theta \left( 1 \right)$  \\
\ \ 4:\textbf{Initialize} \  n = 1,N = {max\_iteration\_count}   & $\Theta \left( {\left( {\left| I \right| + \left| J \right| + \left| K \right|} \right) \times R} \right)$  \\
\ \ 5:\textbf{while}\  $t < T$\ $\textbf{and not}$ converge $\textbf{do}$ &  $ \times {\rm{t}}$\\
\ \ 6:\quad \  \textbf{for} \  $y_{ijk}$ \  $in$ \  $\Lambda $ \textbf{do} &  $ \times \Theta \left( {\left| \Lambda  \right|} \right)$\\
\ \ 7:\quad \ \quad \  $\hat{y}^{t}_{ijk}=\sum_{r=1}^{R} \varphi\left(u^{t}_{ir} \right)\varphi\left(o^{t}_{jr} \right) \varphi\left(m^{t}_{kr} \right)$  &   $ \times\Theta \left(R\right)$\\
\ \ 8:\quad \ \quad \ \textbf{for} \  r=1 \  \textbf{to} \  R \  \textbf{do} &   $\times R$\\
\ \ 9:\quad \ \quad \ \quad \   $u_{ir}^{t + 1} = u_{ir}^t - \left( {\eta {{\partial f_{ijk}^t} \mathord{\left/
 {\vphantom {{\partial f_{ijk}^t} {\partial u_{ir}^t}}} \right.
 \kern-\nulldelimiterspace} {\partial u_{ir}^t}} + {C_I}I_{{u_{ir}}}^{t - 1} + {C_D}\left( {{{\partial f_{ijk}^t} \mathord{\left/
 {\vphantom {{\partial f_{ijk}^t} {\partial u_{ir}^t}}} \right.
 \kern-\nulldelimiterspace} {\partial u_{ir}^t}} - D_{{u_{ir}}}^{t - 1}} \right)} \right)$ &   $\Theta (1)$\\
10:\quad \ \quad \ \quad \ $o_{jr}^{t + 1} = o_{jr}^t - \left( {\eta {{\partial f_{ijk}^t} \mathord{\left/
 {\vphantom {{\partial f_{ijk}^t} {\partial o_{jr}^t}}} \right.
 \kern-\nulldelimiterspace} {\partial o_{jr}^t}} + {C_I}I_{{o_{jr}}}^{t - 1} + {C_D}\left( {{{\partial f_{ijk}^t} \mathord{\left/
 {\vphantom {{\partial f_{ijk}^t} {\partial o_{jr}^t}}} \right.
 \kern-\nulldelimiterspace} {\partial o_{jr}^t}} - D_{{o_{jr}}}^{t - 1}} \right)} \right)$ &   $\Theta (1)$\\
11:\quad \ \quad \ \quad \   $m_{kr}^{t + 1} = m_{kr}^t - \left( {\eta {{\partial f_{ijk}^t} \mathord{\left/
 {\vphantom {{\partial f_{ijk}^t} {\partial m_{kr}^t}}} \right.
 \kern-\nulldelimiterspace} {\partial m_{kr}^t}} + {C_I}I_{{m_{kr}}}^{t - 1} + {C_D}\left( {{{\partial f_{ijk}^t} \mathord{\left/
 {\vphantom {{\partial f_{ijk}^t} {\partial m_{kr}^t}}} \right.
 \kern-\nulldelimiterspace} {\partial m_{kr}^t}} - D_{{m_{kr}}}^{t - 1}} \right)} \right)$ &   $\Theta (1)$\\

 12:\quad \ \quad \ \quad \   $I_{{u_{ir}}}^t = (1 - \alpha )I_{{u_{ir}}}^{t - 1} + \alpha {{\partial f_{ijk}^t} \mathord{\left/
 {\vphantom {{\partial f_{ijk}^t} {\partial u_{ir}^t}}} \right.
 \kern-\nulldelimiterspace} {\partial u_{ir}^t}}$ & $\Theta (1)$\\

 13:\quad \ \quad \ \quad \   $I_{{o_{jr}}}^t = (1 - \alpha )I_{{o_{jr}}}^{t - 1} + {{\alpha \partial f_{ijk}^t} \mathord{\left/
 {\vphantom {{\alpha \partial f_{ijk}^t} {\partial o_{jr}^t}}} \right.
 \kern-\nulldelimiterspace} {\partial o_{jr}^t}}$ & $\Theta (1)$\\
  14:\quad \ \quad \ \quad \   $I_{{m_{kr}}}^t = (1 - \alpha )I_{{m_{kr}}}^{t - 1} + {{\alpha \partial f_{ijk}^t} \mathord{\left/
 {\vphantom {{\alpha \partial f_{ijk}^t} {\partial m_{kr}^t}}} \right.
 \kern-\nulldelimiterspace} {\partial m_{kr}^t}}$ & $\Theta (1)$\\

 15:\quad \ \quad \ \quad \   $D_{{u_{ir}}}^t = {{\partial f_{ijk}^t} \mathord{\left/
 {\vphantom {{\partial f_{ijk}^t} {\partial u_{ir}^t}}} \right.
 \kern-\nulldelimiterspace} {\partial u_{ir}^t}}$ & $\Theta (1)$\\

  16:\quad \ \quad \ \quad \   $D_{{o_{jr}}}^t = {{\partial f_{ijk}^t} \mathord{\left/
 {\vphantom {{\partial f_{ijk}^t} {\partial o_{jr}^t}}} \right.
 \kern-\nulldelimiterspace} {\partial o_{jr}^t}}$ & $\Theta (1)$\\

17:\quad \ \quad \ \quad \   $D_{{m_{kr}}}^t = {{\partial f_{ijk}^t} \mathord{\left/
 {\vphantom {{\partial f_{ijk}^t} {\partial m_{kr}^t}}} \right.
 \kern-\nulldelimiterspace} {\partial m_{kr}^t}}$ & $\Theta (1)$\\
18:\quad \ \quad \ \textbf{end for} & $ - $\\
19:\quad \  \textbf{end for} & $ - $\\
20:\textbf{end while} & $ - $\\
\bottomrule[2pt] 
\end{tabularx}
\end{table*}

\subsection{Algorithm Design and Analysis}

Based on the above inferences, the algorithm \textbf{PNLF} is given. According to Algorithm \textbf{PNLF}, the primary tasks in each iteration include updating LFs and storing their historical updates. Therefore, its computational cost is:
\begin{equation}
    C = \Theta \left( {2 \times t \times R \times \left| \Lambda  \right|} \right) \approx \Theta \left( {t \times R \times \left| \Lambda  \right|} \right).
    \label{eq17}
\end{equation}

Note that in practical scenarios, $\left| \Lambda  \right| \gg \max \left\{ {\left| I \right|,\left| J \right|,\left| K \right|} \right\}$ , as shown in Table~\ref{tab2}, which allows the derivation of (\ref{eq17}). Given that $n$ and $R$ is positive, the computational cost of PNLF model is linear with $\left| \Lambda  \right|$.

Model's storage cost is primarily determined by three factors: 1) LFs; 2) Auxiliary arrays ${I_{(U)}}$, ${I_{(O)}}$, and ${I_{(M)}}$ store the accumulated historical updates of LFs, while ${D_{(U)}}$,  ${D_{(O)}}$, and ${D_{(M)}}$ store the previous updates of LFs; 3) Entries in $\bm{\mathit{Y}}$ and  $\bm{\mathit{\hat Y}}$ corresponding to $\left| \Lambda  \right|$. Therefore, the storage cost of PNLF model is given as:
\begin{align}
    S &= \Theta \left( {3 \times R \times \left( {\left| I \right| + \left| J \right| + \left| K \right|} \right) + 2 \times \left| \Lambda  \right|} \right) \nonumber \\
    &\approx \Theta \left( {R \times \left( {\left| I \right| + \left| J \right| + \left| K \right|} \right) + 2 \times \left| \Lambda  \right|} \right).
\end{align}

From the above, it can be inferred that the storage complexity of the PNLF model is linear with the number of known tensor entries and its LFs.

section{Experimental Results and Analysis}
\label{sec4}

\subsection{Experimental Setup and Evaluation}
\subsubsection{Dataset}

\begin{table*}[htbp]
\caption{Dataset Details}
\newcolumntype{C}{>{\centering\arraybackslash}X}
\begin{tabularx}{\textwidth}{CCCC}
\toprule 
\textbf{Attributes} & \textbf{D1 (iAWE)} & \textbf{D2 (UK-DALE)} & \textbf{D3 (REDD)} \\
\midrule 
Daily Samples & $86400$ & $86400$ & $86400$ \\
Device Quantity & $13$ & $7$ & $9$ \\
Days & $21$ & $21$ & $21$ \\
Known Density & $6.65\times10^{10^{-2}}$ & $3.25\times10^{10^{-2}}$ & $4.80\times10^{10^{-2}}$ \\
Known Count & $1569491$ & $413357$ & $1655421$ \\

\bottomrule 
\end{tabularx}
\label{tab2}
\end{table*}

The experiments in this paper utilized three publicly available datasets: iAWE, REDD \cite{paper31}, and UK-DALE \cite{paper32}. These datasets, which record electricity consumption from various regions and buildings in the real world, were selected due to their inherent data missingness. As shown in Table~\ref{tab2}, sampled data over a period of time were selected for experimentation, with a sampling frequency of 1 Hz and a duration of 21 days.

To ensure stability in the updating process, we conducted linear feature scaling on each dataset, mapping the values to the range [0,10]. The scaling formula is as follows:

\begin{equation}
    {\rm{\tilde y}} = 10 \times \frac{{{\rm{y}} - {{\rm{y}}_{\min }}}}{{{{\rm{y}}_{\max }} - {{\rm{y}}_{\min }}}}
\end{equation}

The accuracy of NILM data interpolation reflects the model's ability to capture the essential characteristics of incomplete tensors. Root Mean Squared Error (RMSE) and Mean Absolute Error (MAE) are commonly used to measure the proximity between target values and estimated values \cite{paper7,paper8,paper9,paper10,paper11,paper12,paper33}. Therefore, RMSE and MAE have been chosen as evaluation metrics. Formally,

\begin{align}
    &RMSE = \sqrt {\frac{{\sum\nolimits_{{y_{ijk}} \in \Omega } {{{\left( {{y_{ijk}} - {{\hat y}_{ijk}}} \right)}^2}} }}{{\left| \Omega  \right|}}} ,
    \nonumber  \\
    &MAE = \frac{{\sum\nolimits_{{y_{ijk}} \in \Omega } {\left| {{y_{ijk}} - {{\hat y}_{ijk}}} \right|} }}{{\left| \Omega  \right|}}.
\end{align}

\subsubsection{General Settings}

In this study, each dataset's $\Phi $ ,  $\Psi $ and $\Omega $  are all mutually exclusive. The training set is utilized for model training, the validation set is used to determine model convergence, and the test set is employed to evaluate model performance. To ensure the objectivity of our results, the following measures were implemented:

a) To enhance the credibility of our experimental outcomes, 20 repetitions of the experiments were conduct-ed to mitigate the impact of random errors in the data.

b) For fair comparisons, the LF space dimension $R$ of the LFT model used in the experiments was set to be the same.

c) The training process was terminated under the following conditions: when the difference between consecutive validation errors fell below 1e-6 (indicating model convergence) or when the number of iterations exceeded the predefined threshold of 200.

\begin{figure*}[!!htbp]
\centering
\includegraphics[width=1 \textwidth]{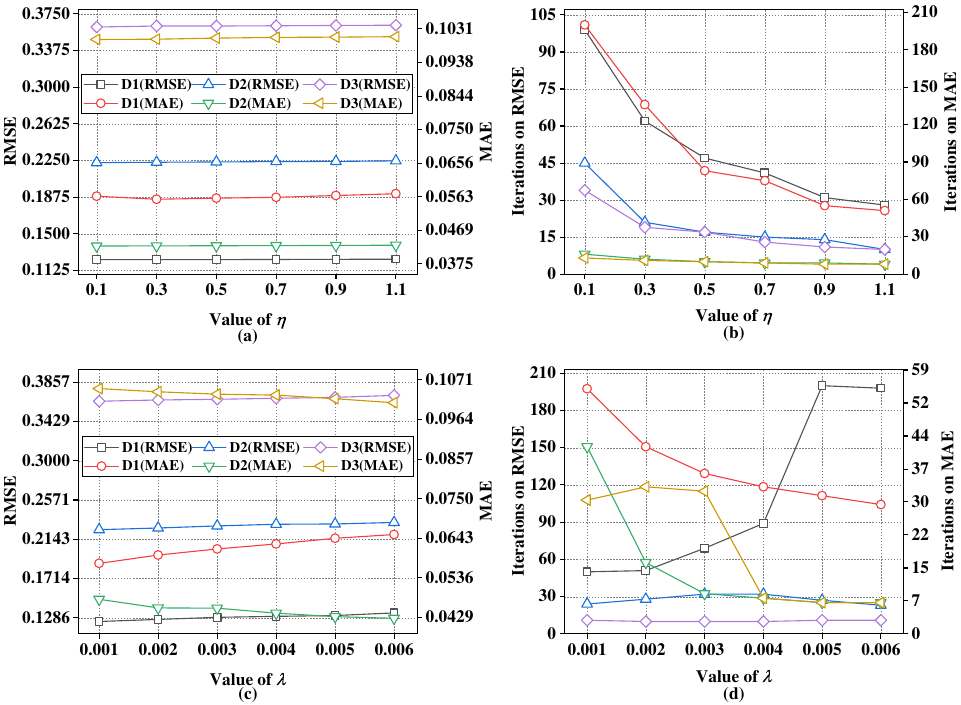}
\caption{Impacts of  $\eta$ and $\lambda$  .\label{fig4}}
\end{figure*} 

\begin{figure*}[!!htbp]
\centering
\includegraphics[width=1 \textwidth]{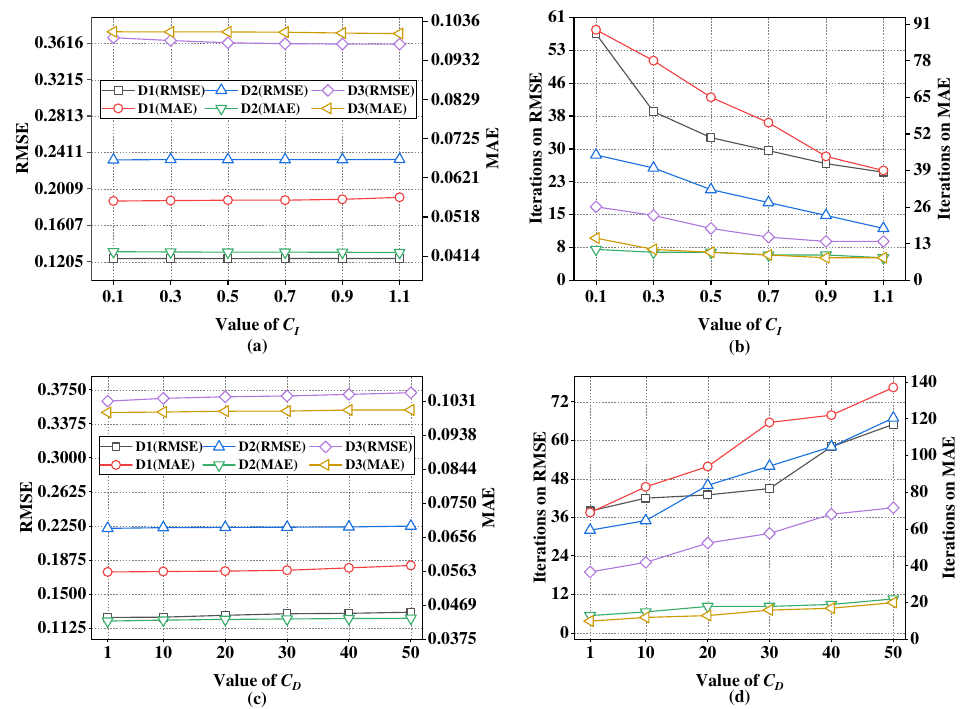}
\caption{Impacts of  $C_{I}$ and $C_{D}$  .\label{fig5}}
\end{figure*}  

\subsection{Parameter Sensitivity Tests}

Based on the analysis in Section 3, the model's performance is significantly influenced by the regularization coefficient $\lambda $, learning rate $\eta$,  and the PID control coefficients ${C_I}$, and  ${C_D}$. Therefore, a sensitivity analysis of these parameters is conducted in this section.

\begin{figure*}[!ht]
\centering
\includegraphics[width=1 \textwidth]{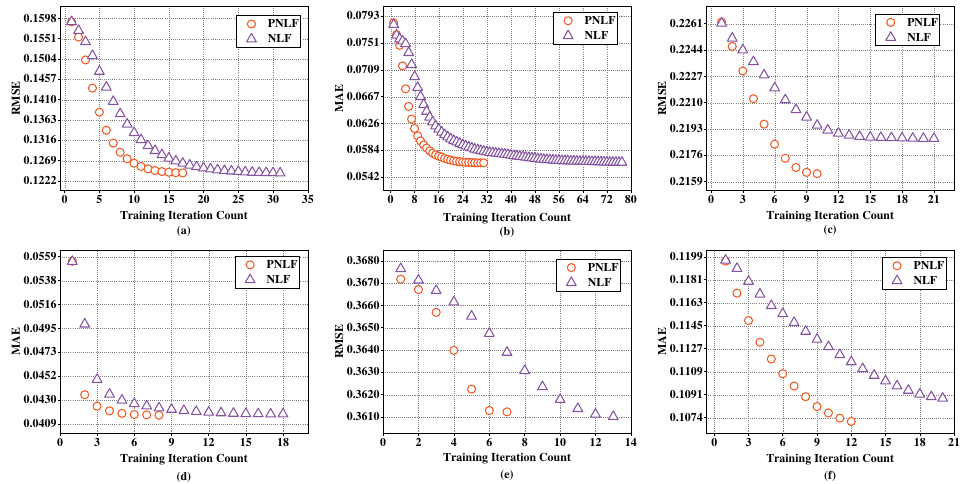}
\caption{Performance of PNLF vs. NLF on D1-D3. \label{fig6}}
\end{figure*}  

\textit{1)	Effects of  $\eta$ and  $\lambda$ }

With other parameters fixed, $\eta $  was increased from 0.1 to 1.1, and $\lambda$ was increased from 0.001 to 0.006. Figures~\ref{fig4}(a) and ~\ref{fig4}(c) present the effects of $\eta $ and $\lambda$ on RMSE and MAE, respectively, while Figure~\ref{fig4}(b) and ~\ref{fig4}(d) illustrate their impact on iteration counts for RMSE and MAE. Based on these results:

\textit{a) The iteration count of the model increases with higher $\eta$,  accompanied by a slight rise in repair error.} As shown in Figure~\ref{fig4}(a), when $\eta $  increases from 0.1 to 1.1 on the D1, the RMSE iteration count decreases from 99 to 28, indicating that a higher $\eta $  value may lead to faster convergence. However, the RMSE value slightly increases from 0.1236 to 0.1241, suggesting that while convergence speed improves, the repair error slightly worsens, potentially due to more aggressive adjustments in the optimization process. Similar trends are observed for MAE and across other datasets, indicating that adjusting $\eta $ can affect both the convergence characteristics and the model's repair accuracy. Therefore, selecting an appropriate $\eta $ value is crucial for balancing convergence speed and repair accuracy.

\textit{b) As $\lambda $ increases, the trends in repair error and iteration counts vary across different datasets and evaluation metrics.} For instance, as shown in Figure~\ref{fig4}(c), on the D1, RMSE increases from 0.1246 to 0.1340 with the rise in $\lambda$, indicating a decline in model performance with larger $\lambda $  values. In contrast, on the D2, MAE decreases from 0.0478 to 0.0427. This indicates that $\lambda $ controls the model's degree of overfitting, and a larger $\lambda $ might enhance model performance in terms of MAE on the D2. Additionally, iteration trends differ as well. For example, as depicted in Figure~\ref{fig4}(d), on the D1, the RMSE iteration count increases from 50 to 198, while the MAE iteration count decreases from 55 to 29 as $\lambda $ increases.

\textit{2)	Effects of  ${C_I}$ and  ${C_D}$ }

In this series of experiments, $\eta$ and $\lambda$  were kept constant while ${C_I}$ and ${C_D}$ were gradually adjusted to study 
their impact on PNLF model's performance. The experimental results are shown in Figure~\ref{fig5}. Based on these results, the following observations were made: 

\textit{a) As ${C_I}$ increases, repair accuracy shows slight variations, while the iteration count decreases progressively}. For instance, as shown in Figure~\ref{fig5}(c), when ${C_I}$ increases from 0.1 to 1.1, RMSE on the D1 slightly rises from 0.1239 to 0.1242, and MAE changes from 0.0560 to 0.0569, indicating minimal impact on accuracy. The effect of ${C_I}$ on repair accuracy varies across different datasets, though the changes are subtle. The most notable effect is the acceleration of model convergence. As illustrated in Figure~\ref{fig5}(d), on the D1, increasing ${C_I}$ from 0.1 to 1.1 results in a significant reduction in RMSE iteration count from 57 to 25, and MAE iteration count drops from 89 to 39, with similar patterns observed across other datasets.

\textit{b) As increases, both repair error and iteration count show a significant upward trend}. For example, as shown in Figures~\ref{fig5}(c) and ~\ref{fig5}(d), when ${C_D}$ increases from 1 to 50, RMSE on the D1 rises from 0.1243 to 0.1300, with 
iteration count increasing from 38 to 65. Similarly, MAE increases from 0.0561 to 0.0579, and iteration count rises from 69 to 137. This pattern is consistently observed across other datasets, indicating that higher ${C_D}$ values not only lead to greater repair errors but also significantly increase the iteration count. This change may result from the model struggling to converge due to the increased complexity caused by excessive adjustment.

\subsection{Ablation Analysis: PNLF Compared to NLF}

To investigate the impact of the PID controller on the Non-negative Latent Factorization of Tensors (NLF) model, an ablation study was conducted comparing two models: one is the PNLF model, and the other is the NLF model without PID control (i.e., with $C_{I}$  = $C_{D}$ = 0 in the PNLF model). After ensuring both models have identical $\lambda$ and  $\eta$ values, the performance was evaluated by adjusting $C_{I}$ and $C_{D}$ in the PNLF model. Figure~\ref{fig6} presents the iterative curves and performance metrics. Based on these results, the following conclusions were drawn:

\textit{The introduction of PID control significantly reduces the iteration count of the NLF model, while the repair accuracy of both models remains nearly identical}. As shown in Figure~\ref{fig6}(a), on the D1, the iteration counts for the NLF and PNLF models are 31 and 17, respectively, with the latter reducing the count by approximately 45.17\% (calculated as (larger value - smaller value) / larger value). The RMSE at convergence for both models is 0.1240 and 0.1241, respectively. Similar results are observed on the D2 and D3. For MAE, as shown in Figure~\ref{fig6}(d), on the D2, the iteration counts for the NLF and PNLF models are 18 and 8, respectively, with the latter reducing the count by about 55.56\%. The MAE at convergence is 0.0418 and 0.0417, respectively, with similar trends ob-served on the D1 and D3.

\subsection{Comparison With State-of-the-Art Models}

Experiments were conducted on a computer equipped with an Intel Core i7-10700 processor (2.9 GHz) and 16 GB RAM. Python 3.11.5 was selected as the primary programming platform. In this section, the PNLF model is compared with several state-of-the-art tensor completion (TC) models to validate its performance. The details of the compared models are as follows:
\begin{enumerate}
\item	\textbf{M1}: A LFT model \cite{paper10} incorporating linear bias, it includes linear bias in the learning objective and employs SGD-based multiplicative update rules to ensure non-negativity;
\item	\textbf{M2}: A classic low-rank TC model \cite{paper34} with high-accuracy completion, based on minimizing the nuclear norm (MNN) and implemented using the Alternating Direction Method of Multipliers (ADMM) algorithm;
\item	\textbf{M3}: A TC model \cite{paper35} based on approximate singular value decomposition, it utilizes QR decomposition to approximate the singular value decomposition process, and enhances model robustness by modeling noise;
\item \textbf{M4}: A TC model \cite{paper36} employing AdamW as the learning scheme, approximating the target tensor through the product of three smaller tensors;
\item \textbf{M5}: A PNLF model proposed in this paper.
\end{enumerate}

With the LF dimension $R$ set to 20 for all LFT models, and when the validation and training sets have the same ratio, with the remainder used as the test set, Figure~\ref{fig7} shows the repair accuracy of all models across different training set proportions. Table~\ref{tab3} presents the repair accuracy of each model under a 6:2:2 split for the training, validation, and test sets, while Tables~\ref{tab4} and ~\ref{tab5} detail their time and storage costs. Figure~\ref{fig8} illustrates the repair performance of the PNLF model under different $R$ values and training set proportions. These results reveal the following insights:

\begin{figure*}[ht]
\centering
\includegraphics[width=1 \textwidth]{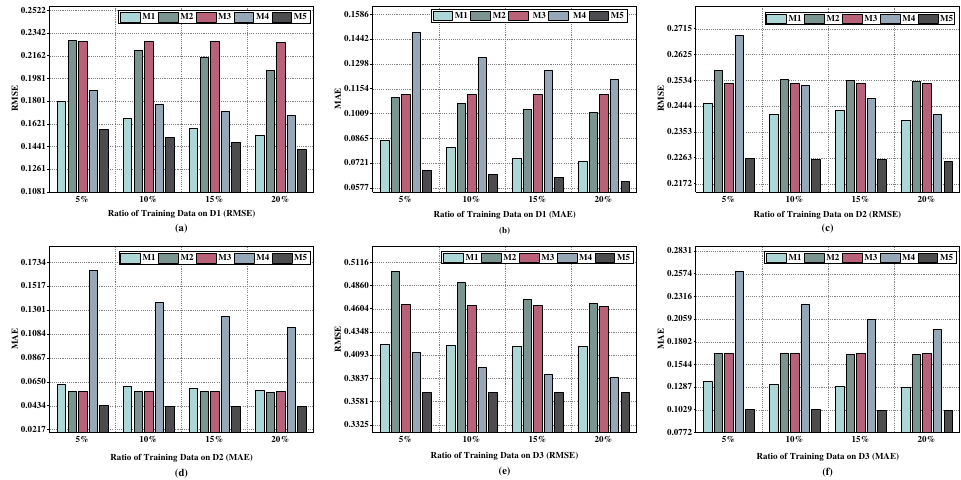}
\caption{RMSE and MAE of  M1-5 on D1-3 with Different Training Ratios.\label{fig7}}
\end{figure*}  

\begin{figure*}[ht]
\centering
\includegraphics[width=1 \textwidth]{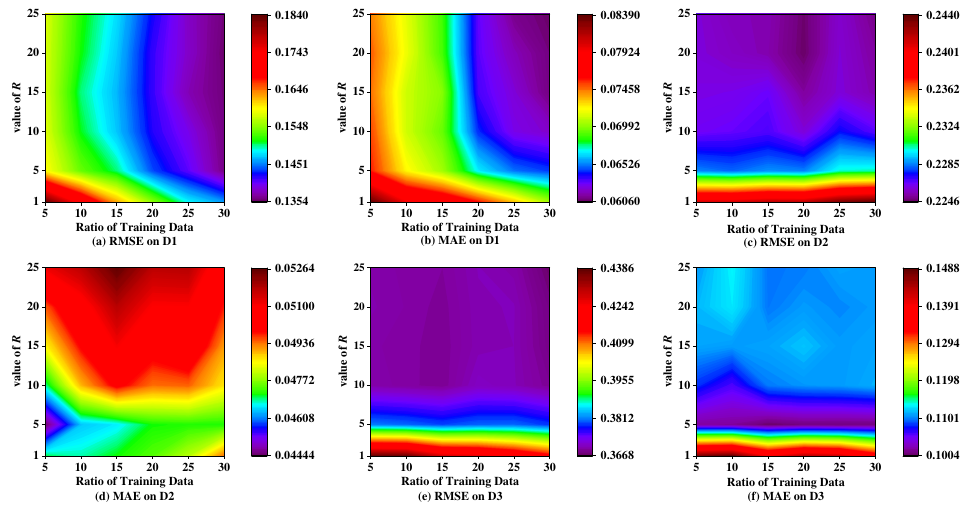}
\caption{Effects of LF Dimension $R$ on PNLF Model Performance (Training Set Ratio Equal to Validation Set, Remainder as Test Set) \label{fig8}}
\end{figure*}

\textit{a) Under extremely sparse data conditions, the PNLF model outperforms all other models:} As shown in Figure~\ref{fig7}, with a 5\% validation set, M5's RMSE on D1-3 is 0.1577, 0.2261, and 0.3690, respectively, representing improvements of 12.63\%, 7.90\%, and 10.48\% over the next best results of 0.1805, 0.2455, and 0.4122. This trend is consistent across RMSE, MAE, and various training set ratios, highlighting the PNLF model's strong adaptability and robustness in handling highly sparse data. It is noteworthy that M2 and M3, based on NNM, perform poorly under such sparse conditions, likely due to the limitations of this approach in extreme sparsity scenarios.

\textit{b) The impact of the LF dimension $R$ on the PNLF model's estimation error is not entirely monotonic, necessitating careful selection of $R$:} As shown in Figure ~\ref{fig8}, the model's accuracy is low when R is small. For instance, in Figure~\ref{fig8}(a), with $R$=1 and a training set proportion of 5\%, the RMSE is 0.1838. When $R$ increases to 25, the RMSE improves to 0.1580, reflecting a 14.04\% enhancement. Conversely, when $R$ becomes too large, its effect on accuracy diminishes. For example, with a training set proportion of 5\%, the RMSE values for $R$=15 and $R$=20 are 0.1579 and 0.1578, respectively. Similar trends are observed under other conditions and in most subplots. An exception is Figure~\ref{fig8}(d), where the lowest MAE is consistently achieved at $R$=5, regardless of the training set proportion, after which MAE increases with rising $R$. For instance, with a training set proportion of 30\%, the MAE values for $R$=5 and $R$=25 are 0.0472 and 0.0504, respectively. It is important to note that increasing $R$ significantly raises time and memory consumption, so the selection of $R$ should balance accuracy with time and resource usage.

\begin{table*}[h] 
\caption{RMSE and MAE of M1-5 on D1-3.\label{tab3}}
\newcolumntype{C}{>{\centering\arraybackslash}X}
\begin{tabularx}{\textwidth}{CCCCCC}
\toprule
\textbf{RMSE}	& \textbf{M1}	& \textbf{M2} & \textbf{M3} & \textbf{M4} & \textbf{M5} \\
\midrule
D1	& ${0.1318_{ \pm 5e - 4}}$ & ${0.1570_{ \pm 0}}$ & ${0.2261_{ \pm 4e - 6}}$ & ${0.1484_{ \pm 4e - 6}}$ & $\bm{0.1238_{ \pm 4e - 4}}$  \\
D2	& ${0.2342_{ \pm 1e - 3}}$ & ${0.2551_{ \pm 0}}$ & ${0.2594_{ \pm 6e - 6}}$ & ${0.2371_{ \pm 7e - 6}}$ & $\bm{0.2239_{ \pm 5e - 4}}$  \\
D3	& ${0.4054_{ \pm 3e - 4}}$ & ${0.4253_{ \pm 0}}$ & ${0.4663_{ \pm 4e - 6}}$ & ${0.3751_{ \pm 2e - 5}}$ & $\bm{0.3631_{ \pm 7e - 4}}$  \\
\midrule
\textbf{MAE}	& \textbf{M1}	& \textbf{M2} & \textbf{M3} & \textbf{M4} & \textbf{M5} \\
\midrule
D1	& ${0.0651_{ \pm 1e - 3}}$ & ${0.0769_{ \pm 0}}$ & ${0.1112_{ \pm 1e - 6}}$ & ${0.0807_{ \pm 6e - 6}}$ & $\bm{0.0566_{ \pm 2e - 4}}$\\
D2	& ${0.0643_{ \pm 3e - 3}}$ & ${0.0541_{ \pm 0}}$ & ${0.0572_{ \pm 7e - 6}}$ & ${0.0871_{ \pm 1e - 5}}$ & $\bm{0.0425_{ \pm 4e - 5}}$\\
D3 	& ${0.1311_{ \pm 1e - 3}}$ & ${0.1497_{ \pm 0}}$ & ${0.1678_{ \pm 7e - 6}}$ & ${0.1651_{ \pm 2e - 5}}$ & $\bm{0.0998_{ \pm 7e - 5}}$\\
\bottomrule
\end{tabularx}
\footnotesize{Note: The bold values represent the optimal values, while \(\pm\) indicates the standard deviation for each corresponding value.}
\end{table*}

\begin{table*}[h] 
\caption{Time Costs (seconds) of M1-5 on D1-3.\label{tab4}}
\newcolumntype{C}{>{\centering\arraybackslash}X}
\begin{tabularx}{\textwidth}{CCCCCC}
\toprule
\textbf{RMSE}	& \textbf{M1}	& \textbf{M2} & \textbf{M3} & \textbf{M4} & \textbf{M5} \\
\midrule
D1	& ${84.8_{ \pm 61.21}}$ & ${337.45_{ \pm 1.11}}$ & ${80.25_{ \pm 1.41}}$ & ${148.3_{ \pm 1.30}}$ & $\bm{24.6_{ \pm 2.33}}$   \\
D2	& ${41.35_{ \pm 3.91}}$ & ${89.55_{ \pm 1.35}}$ & ${36.8_{ \pm 1.67}}$ & ${28.15_{ \pm 2.61}}$ & $\bm{23.4_{ \pm 2.44}}$   \\
D3	& ${13.55_{ \pm 1.95}}$ & ${22.15_{ \pm 0.65}}$ & ${35.7_{ \pm 1.54}}$ & ${7.55_{ \pm 0.97}}$ & $\bm{3.95_{ \pm 1.29}}$   \\
\midrule
\textbf{MAE}	& \textbf{M1}	& \textbf{M2} & \textbf{M3} & \textbf{M4} & \textbf{M5} \\
\midrule
D1	& ${289.15_{ \pm 17.78}}$ & ${1510.4_{ \pm 1.01}}$ & ${72.35_{ \pm 1.47}}$ & ${157.65_{ \pm 1.98}}$ & $\bm{{56.05_{ \pm 4.21}}}$   \\
D2	& ${81.15_{ \pm 76.25}}$ & ${103.2_{ \pm 0.67}}$ & ${41.5_{ \pm 1.69}}$ & ${36.2_{ \pm 1.43}}$ & $\bm{10.35_{ \pm 1.52}}$   \\
D3	& ${46.2_{ \pm 15.96}}$ & ${24.25_{ \pm 0.88}}$ & ${13.2_{ \pm 1.52}}$ & ${10.85_{ \pm 1.24}}$ & $\bm{3.35_{ \pm 1.12}}$   \\
\bottomrule
\end{tabularx}
\end{table*}

\begin{table*}[h] 
\caption{Storage Costs of M1-5 on D1-3.\label{tab5}}
\newcolumntype{C}{>{\centering\arraybackslash}X}
\begin{tabularx}{\textwidth}{CCCCCC}
\toprule
\textbf{Storage Costs (MB)} & \textbf{M1} & \textbf{M2} & \textbf{M3} & \textbf{M4} & \textbf{M5} \\
\midrule
D1 & ${278_{ \pm 12}}$ & ${2069_{ \pm 237}}$ & ${2927_{ \pm 325}}$ & ${820_{ \pm 54}}$ & $\bm{265_{ \pm 26}}$  \\
D2 & ${275_{ \pm 21}}$ & ${2787_{ \pm 572}}$ & ${3704_{ \pm 778}}$ & ${890_{ \pm 84}}$ & $\bm{269_{ \pm 21}}$  \\
D3 & ${242_{ \pm 17}}$ & ${1122_{ \pm 231}}$ & ${1497_{ \pm 325}}$ & ${637_{ \pm 115}}$ & $\bm{228_{ \pm 26}}$  \\
\bottomrule
\end{tabularx}
\end{table*}

\textit{c) The PNLF model outperforms its peers in repair accuracy:} As shown in Table~\ref{tab3}, M5 achieves RMSE values of 0.1238, 0.2239, and 0.3631 on D1-3, respectively, representing improvements of 6.01\%, 4.4\%, and 3.2\% over the next best results of 0.1318, 0.2342, and 0.3751. Similarly, M5's MAE values of 0.0566, 0.0425, and 0.0998 across D1-3 are lower, with improvements of 13.06\%, 21.44\%, and 23.87\% compared to the next best results of 0.0651, 0.0541, and 0.1311. These results indicate PNLF model's consistent advantage, particularly in han-dling complex or noisy data.

\textit{d) In terms of computational efficiency, PNLF model exhibits a clear advantage over other models:} As shown in Table~\ref{tab4}, the RMSE convergence times for M5 are 24.6, 23.4, and 3.95 seconds for D1-3, respectively. Compared to the second-best results, with convergence times of 80.25, 28.15, and 7.55 seconds, M5's times are 30.65\%, 83.12\%, and 52.32\% of these times. For MAE, M5's convergence times are 56.05, 10.35, and 3.35 seconds for D1-3, respectively. These are 77.47\%, 28.59\%, and 30.88\% of the second-best times, which are 72.35, 36.2, and 10.85 seconds, respectively.

\textit{e) PNLF model's storage costs are competitive and fall within a reasonable range compared to its peers:} According to Table~\ref{tab5}, M5 has the lowest storage costs among all compared models in D1-3, with values of 265, 269, and 214, respectively. The closest competitors are M1, with storage costs of 278, 275, and 242 in D1-3. Notably, models based on MNN, such as M2 and M3, have higher storage costs due to the involvement of singular value matrices and the need to produce full tensors during the update process.

\subsection{Summary}

Based on the experimental results, the advantages of the PNLF model are highlighted: a) fewer iterations com-pared to the LFT model, b) higher computational efficiency, c) highly competitive repair accuracy, and d) man-ageable storage costs. Therefore, the PNLF model is better suited for handling incomplete NILM data.

\section{Discussion}
\label{sec5}

In this study, we proposed a novel Proportional-Integral-Derivative (PID) Controlled Non-Negative Latent Factorization of Tensor (PNLF) model to address missing data in Non-Intrusive Load Monitoring (NILM) and improve load disaggregation accuracy. The PNLF model demonstrates the following advantages: a) The use of the Sigmoid function ensures the non-negativity of the data, enhancing the model’s ability to handle nonlinear relationships; b) The integration of a PID controller dynamically adjusts gradient increments, resulting in faster convergence and greater stability.

However, despite these encouraging results, there are several aspects that require further discussion and improvement:

\textbf{Model Complexity and PID Hyperparameter Tuning:}
While the introduction of the PID controller significantly improves convergence speed and stability, it also introduces additional hyperparameters that require careful tuning. The manual grid search process for adjusting these parameters can be time-consuming. Future research could explore adaptive hyperparameter tuning techniques, such as Bayesian optimization \cite{paper60} or evolutionary algorithms \cite{paper40}, to reduce the need for manual tuning and improve efficiency.

\textbf{Limitations of Shallow Tensor Factorization:}
Our model, like many tensor factorization methods, employs relatively shallow factorization techniques. While effective for capturing the primary structure of the data, deep neural networks have been shown to more effectively capture hidden and complex patterns in high-dimensional data \cite{paper37}. Recent advancements in deep matrix factorization \cite{paper38} suggest that extending tensor factorization to deep architectures could enhance the ability to capture more intricate relationships in NILM data. Investigating the potential of deep tensor decomposition for handling missing data in NILM is a promising direction for future research.

\textbf{Scalability and Computational Efficiency:}
Although the PNLF model demonstrates competitive accuracy and convergence speed on three real NILM datasets, its computational efficiency when handling large-scale datasets or real-time applications remains an open question. Future research should focus on optimizing the scalability of the model, possibly by utilizing distributed computing frameworks or reducing memory consumption.

\bibliographystyle{ieeetr}
\bibliography{paper}

\begin{thebibliography}{10}

\bibitem{paper41}
H.~Yue, K.~Yan, J.~Zhao, Y.~Ren, X.~Yan, and H.~Zhao, ``Estimating demand response flexibility of smart home appliances via nilm algorithm,'' in {\em 2020 IEEE 4th Information Technology, Networking, Electronic and Automation Control Conference (ITNEC)}, vol.~1, pp.~394--398, IEEE, 2020.

\bibitem{paper42}
J.~M. Alcala, J.~Urena, A.~Hernandez, and D.~Gualda, ``Sustainable homecare monitoring system by sensing electricity data,'' {\em IEEE Sensors Journal}, vol.~17, no.~23, pp.~7741--7749, 2017.

\bibitem{paper43}
F.~Kalinke, P.~Bielski, S.~Singh, E.~Fouch{\'e}, and K.~B{\"o}hm, ``An evaluation of nilm approaches on industrial energy-consumption data,'' in {\em Proceedings of the twelfth ACM international conference on future energy systems}, pp.~239--243, 2021.

\bibitem{paper44}
G.-F. Angelis, C.~Timplalexis, S.~Krinidis, D.~Ioannidis, and D.~Tzovaras, ``Nilm applications: Literature review of learning approaches, recent developments and challenges,'' {\em Energy and Buildings}, vol.~261, p.~111951, 2022.

\bibitem{paper45}
A.~Allik and A.~Annuk, ``Interpolation of intra-hourly electricity consumption and production data,'' in {\em 2017 IEEE 6th international conference on renewable energy research and applications (ICRERA)}, pp.~131--136, IEEE, 2017.

\bibitem{paper46}
X.~Miao, Y.~Gao, G.~Chen, B.~Zheng, and H.~Cui, ``Processing incomplete k nearest neighbor search,'' {\em IEEE Transactions on Fuzzy Systems}, vol.~24, no.~6, pp.~1349--1363, 2016.

\bibitem{paper47}
P.~Royston, ``Multiple imputation of missing values,'' {\em The Stata Journal}, vol.~4, no.~3, pp.~227--241, 2004.

\bibitem{paper48}
S.~Dray and J.~Josse, ``Principal component analysis with missing values: a comparative survey of methods,'' {\em Plant Ecology}, vol.~216, pp.~657--667, 2015.

\bibitem{paper49}
Y.~Koren, ``Collaborative filtering with temporal dynamics,'' in {\em Proceedings of the 15th ACM SIGKDD international conference on Knowledge discovery and data mining}, pp.~447--456, 2009.

\bibitem{paper50}
D.~Wu, Y.~He, X.~Luo, and M.~Zhou, ``A latent factor analysis-based approach to online sparse streaming feature selection,'' {\em IEEE Transactions on Systems, Man, and Cybernetics: Systems}, vol.~52, no.~11, pp.~6744--6758, 2021.

\bibitem{paper51}
J.~Liu, P.~Musialski, P.~Wonka, and J.~Ye, ``Tensor completion for estimating missing values in visual data,'' {\em IEEE transactions on pattern analysis and machine intelligence}, vol.~35, no.~1, pp.~208--220, 2012.

\bibitem{paper52}
M.~E. Kilmer and C.~D. Martin, ``Factorization strategies for third-order tensors,'' {\em Linear Algebra and its Applications}, vol.~435, no.~3, pp.~641--658, 2011.

\bibitem{paper53}
O.~Semerci, N.~Hao, M.~E. Kilmer, and E.~L. Miller, ``Tensor-based formulation and nuclear norm regularization for multienergy computed tomography,'' {\em IEEE Transactions on Image Processing}, vol.~23, no.~4, pp.~1678--1693, 2014.

\bibitem{paper54}
Z.~Zhang, G.~Ely, S.~Aeron, N.~Hao, and M.~Kilmer, ``Novel methods for multilinear data completion and de-noising based on tensor-svd,'' in {\em Proceedings of the IEEE conference on computer vision and pattern recognition}, pp.~3842--3849, 2014.

\bibitem{paper55}
Z.~Zhang and S.~Aeron, ``Exact tensor completion using t-svd,'' {\em IEEE Transactions on Signal Processing}, vol.~65, no.~6, pp.~1511--1526, 2016.

\bibitem{paper56}
Q.~Song, H.~Ge, J.~Caverlee, and X.~Hu, ``Tensor completion algorithms in big data analytics,'' {\em ACM Transactions on Knowledge Discovery from Data (TKDD)}, vol.~13, no.~1, pp.~1--48, 2019.

\bibitem{paper10}
X.~Luo, H.~Wu, H.~Yuan, and M.~Zhou, ``Temporal pattern-aware qos prediction via biased non-negative latent factorization of tensors,'' {\em IEEE transactions on cybernetics}, vol.~50, no.~5, pp.~1798--1809, 2019.

\bibitem{paper57}
E.~Acar, D.~M. Dunlavy, T.~G. Kolda, and M.~M{\o}rup, ``Scalable tensor factorizations for incomplete data,'' {\em Chemometrics and Intelligent Laboratory Systems}, vol.~106, no.~1, pp.~41--56, 2011.

\bibitem{paper58}
W.~Zhang, H.~Sun, X.~Liu, and X.~Guo, ``Temporal qos-aware web service recommendation via non-negative tensor factorization,'' in {\em Proceedings of the 23rd international conference on World wide web}, pp.~585--596, 2014.

\bibitem{paper59}
Y.~Wu, H.~Tan, Y.~Li, J.~Zhang, and X.~Chen, ``A fused cp factorization method for incomplete tensors,'' {\em IEEE transactions on neural networks and learning systems}, vol.~30, no.~3, pp.~751--764, 2018.

\bibitem{paper12}
H.~Wu, X.~Luo, M.~Zhou, M.~J. Rawa, K.~Sedraoui, and A.~Albeshri, ``A pid-incorporated latent factorization of tensors approach to dynamically weighted directed network analysis,'' {\em IEEE/CAA Journal of Automatica Sinica}, vol.~9, no.~3, pp.~533--546, 2021.

\bibitem{paper30}
N.~Batra, M.~Gulati, A.~Singh, and M.~B. Srivastava, ``It's different: Insights into home energy consumption in india,'' in {\em Proceedings of the 5th ACM Workshop on Embedded Systems For Energy-Efficient Buildings}, pp.~1--8, 2013.

\bibitem{paper23}
R.~A. Harshman {\em et~al.}, ``Foundations of the parafac procedure: Models and conditions for an “explanatory” multi-modal factor analysis,'' {\em UCLA working papers in phonetics}, vol.~16, no.~1, p.~84, 1970.

\bibitem{paper11}
Q.~Wang, M.~Chen, M.~Shang, and X.~Luo, ``A momentum-incorporated latent factorization of tensors model for temporal-aware qos missing data prediction,'' {\em Neurocomputing}, vol.~367, pp.~299--307, 2019.

\bibitem{paper24}
H.~Ma, D.~Zhou, C.~Liu, M.~R. Lyu, and I.~King, ``Recommender systems with social regularization,'' in {\em Proceedings of the fourth ACM international conference on Web search and data mining}, pp.~287--296, 2011.

\bibitem{paper9}
J.~Li, X.~Luo, Y.~Yuan, and S.~Gao, ``A nonlinear pid-incorporated adaptive stochastic gradient descent algorithm for latent factor analysis,'' {\em IEEE Transactions on Automation Science and Engineering}, 2023.

\bibitem{paper25}
K.~H. Ang, G.~Chong, and Y.~Li, ``Pid control system analysis, design, and technology,'' {\em IEEE transactions on control systems technology}, vol.~13, no.~4, pp.~559--576, 2005.

\bibitem{paper26}
W.~An, H.~Wang, Q.~Sun, J.~Xu, Q.~Dai, and L.~Zhang, ``A pid controller approach for stochastic optimization of deep networks,'' in {\em Proceedings of the IEEE conference on computer vision and pattern recognition}, pp.~8522--8531, 2018.

\bibitem{paper27}
S.~Ghosh, A.~Dasgupta, and A.~Swetapadma, ``A study on support vector machine based linear and non-linear pattern classification,'' in {\em 2019 International Conference on Intelligent Sustainable Systems (ICISS)}, pp.~24--28, IEEE, 2019.

\bibitem{paper28}
X.~Luo, H.~Wu, and Z.~Li, ``Neulft: A novel approach to nonlinear canonical polyadic decomposition on high-dimensional incomplete tensors,'' {\em IEEE Transactions on Knowledge and Data Engineering}, vol.~35, no.~6, pp.~6148--6166, 2022.

\bibitem{paper7}
Y.~Yuan, X.~Luo, and M.-S. Shang, ``Effects of preprocessing and training biases in latent factor models for recommender systems,'' {\em Neurocomputing}, vol.~275, pp.~2019--2030, 2018.

\bibitem{paper8}
D.~Wu, M.~Shang, X.~Luo, and Z.~Wang, ``An l 1-and-l 2-norm-oriented latent factor model for recommender systems,'' {\em IEEE Transactions on Neural Networks and Learning Systems}, vol.~33, no.~10, pp.~5775--5788, 2021.

\bibitem{paper29}
M.~Hardt, B.~Recht, and Y.~Singer, ``Train faster, generalize better: Stability of stochastic gradient descent,'' in {\em International conference on machine learning}, pp.~1225--1234, PMLR, 2016.

\bibitem{paper39}
W.~Wu, X.~Jing, W.~Du, and G.~Chen, ``Learning dynamics of gradient descent optimization in deep neural networks,'' {\em Science China Information Sciences}, vol.~64, pp.~1--15, 2021.

\bibitem{paper31}
M.~J. Johnson and J.~Z. Kolter, ``A public data set for energy disaggregation research,'' {\em Data Mining Applications in Sustainability}, 2011.

\bibitem{paper32}
J.~Kelly and W.~Knottenbelt, ``The uk-dale dataset, domestic appliance-level electricity demand and whole-house demand from five uk homes,'' {\em Scientific data}, vol.~2, no.~1, pp.~1--14, 2015.

\bibitem{paper33}
F.~Zhang, T.~Gong, V.~E. Lee, G.~Zhao, C.~Rong, and G.~Qu, ``Fast algorithms to evaluate collaborative filtering recommender systems,'' {\em Knowledge-Based Systems}, vol.~96, pp.~96--103, 2016.

\bibitem{paper34}
J.~Liu, P.~Musialski, P.~Wonka, and J.~Ye, ``Tensor completion for estimating missing values in visual data,'' {\em IEEE transactions on pattern analysis and machine intelligence}, vol.~35, no.~1, pp.~208--220, 2012.

\bibitem{paper35}
F.~Wu, C.~Li, Y.~Li, and N.~Tang, ``Robust low-rank tensor completion via new regularized model with approximate svd,'' {\em Information Sciences}, vol.~629, pp.~646--666, 2023.

\bibitem{paper36}
H.~Chen, M.~Lin, J.~Liu, H.~Yang, C.~Zhang, and Z.~Xu, ``Nt-dptc: a non-negative temporal dimension preserved tensor completion model for missing traffic data imputation,'' {\em Information Sciences}, vol.~653, p.~119797, 2024.

\bibitem{paper60}
J.~Wu, X.-Y. Chen, H.~Zhang, L.-D. Xiong, H.~Lei, and S.-H. Deng, ``Hyperparameter optimization for machine learning models based on bayesian optimization,'' {\em Journal of Electronic Science and Technology}, vol.~17, no.~1, pp.~26--40, 2019.

\bibitem{paper40}
S.~B. Joseph, E.~G. Dada, A.~Abidemi, D.~O. Oyewola, and B.~M. Khammas, ``Metaheuristic algorithms for pid controller parameters tuning: Review, approaches and open problems,'' {\em Heliyon}, vol.~8, no.~5, 2022.

\bibitem{paper37}
A.~Krizhevsky, I.~Sutskever, and G.~E. Hinton, ``Imagenet classification with deep convolutional neural networks,'' {\em Advances in neural information processing systems}, vol.~25, 2012.

\bibitem{paper38}
S.~Arora, N.~Cohen, W.~Hu, and Y.~Luo, ``Implicit regularization in deep matrix factorization,'' {\em Advances in Neural Information Processing Systems}, vol.~32, 2019.

\end{thebibliography}

\end{document}